%% file: naaclhlt2019.tex
\documentclass[11pt,a4paper]{article}
\usepackage[hyperref]{acl2019}
\usepackage{times}
\usepackage{latexsym}
\usepackage{amssymb}
\usepackage{amsmath}
\usepackage{bm}
\usepackage{booktabs}
\usepackage{subcaption}
\usepackage{caption}

\usepackage{url}
\usepackage{graphicx}

\usepackage{algorithm}
\usepackage{algorithmic}

\aclfinalcopy 

\title{Learning to Explain: Answering Why-Questions via Rephrasing} 

\author{Allen Nie$^1$ ~~ Erin D. Bennett$^{2}$ ~~ Noah D. Goodman$^{1,2}$ \\
  $^1$Department of Computer Science ~~ $^2$Department of Psychology \\
  Stanford University \\
  {\small \tt anie@cs.stanford.edu ~~ \{erindb,ngoodman\}@stanford.edu} \\
}

\date{}

\begin{document}
\maketitle
\begin{abstract}
Providing plausible responses to why questions is a challenging but critical goal for language based human-machine interaction. 
Explanations are challenging in that they require many different forms of abstract knowledge and reasoning.
Previous work has either relied on human-curated structured knowledge bases or detailed domain representation to generate satisfactory explanations. They are also often limited to ranking pre-existing explanation choices.
In our work, we contribute to the under-explored area of generating natural language explanations for general phenomena.
We automatically collect large datasets of explanation-phenomenon pairs which allow us to train sequence-to-sequence models to generate natural language explanations.
We compare different training strategies and evaluate their performance using both automatic scores and human ratings.
We demonstrate that our strategy is sufficient to generate highly plausible explanations for general open-domain phenomena compared to other models trained on different datasets.
\end{abstract}




\input{intro_short.tex}
\input{task.tex}

\input{model_new.tex}

\input{data.tex}

\input{experiments.tex}

\input{discussion.tex}

\input{conclusion.tex}

\bibliography{naaclhlt2019}
\bibliographystyle{acl_natbib}


\setcounter{figure}{0}
\renewcommand{\thefigure}{S\arabic{figure}}
\setcounter{table}{0}  
\renewcommand{\thetable}{S\arabic{table}}
\appendix

\input{appendix.tex}

\end{document}

%% file: intro_short.tex
\section{Introduction}


Allowing machines to provide human acceptable explanations has long been a difficult task for natural language interaction~\cite{carenini1993generating}. In order to provide explanations, systems need to acquire sophisticated domain-knowledge~\cite{winograd1971procedures}, conduct causal reasoning over complex set of events~\cite{hesslow1988problem} and over narrative chains~\cite{chambers2008unsupervised}, and apply commonsense knowledge~\cite{levesque2011winograd}.



Past work has demonstrated that by leveraging human-curated structured knowledge bases such as WordNet~\cite{miller1995wordnet} or ConceptNet~\cite{liu2004conceptnet}, a system can learn to rank or choose between multiple plausible explanations and reach high accuracy~\cite{luo2016commonsense,sasaki2017handling}. Recent successes have also shown that structured knowledge is not needed if one can train a language model on a large quantity of text. Such model can rank explanations based on the probability that each explanation might appear in natural text~\cite{trinh2018simple}.

\begin{figure}
    \centering
\noindent\fbox{%
    \begin{minipage}{\linewidth}
          \textbf{Phenomenon} The city councilmen refused the demonstrators a permit because $\rule{1cm}{0.15mm}$ ? \\
          \textbf{Original} The city councilmen feared violence.
    \end{minipage}
}
\noindent\fbox{%
    \begin{minipage}{\linewidth}
    \textbf{L2E-Seq2Seq (greedy)}: \\ 
    They were not allowed to march in the city. \\
    \textbf{L2E-Seq2Seq (beam)}: \\
    They did not have a permit. \\
    \textbf{LM-1B}: 
    They were not allowed to use the Cape Town airport. \\
    \textbf{L2W}:
    It was the only thing in the city that could be done.\\
    \textbf{Open-Subtitle}: 
    I don't know.
    \end{minipage}
}
    \caption{We show the original Winograd schema sentence, the original offered explanation, and generated responses from our models.}
    \label{fig:my_label}
\end{figure}


While ranking explanations is an important task, the nature of explanation is more general than this. 
For one phenomenon, there might be many acceptable, natural, and useful explanations.
In our work, instead of simply ranking or choosing explanations generated by humans, we propose to advance this important domain by directly generating the explanation. We measure success based on whether the generated sequence is grammatically correct and is a fluent, natural, and plausible explanation. This task has two advantages. First, it allows us to explore whether such a task is computationally feasible given the current learning framework. Second, answering open-domain why-questions with plausible answers can make chitchat dialogue system more engaging, especially in response to ``why'' questions (which previous systems typically answer with degenerate responses such as ``I don't know'').

We show that simply training a language model on previously existing datasets is not enough. However,
by leveraging dependency parsing patterns, we are able to construct two new datasets that will allow modern neural networks to learn to generate general-domain explanations plausible to humans.
These new datasets of naturally occurring self-explanations (statements with ``because'', unprompted by a question) provide excellent training signal for generating novel explanations for a given phenomenon.
We conduct human experiments on the important features that contribute to plausible explanations, and we describe a simple procedure that can rephrase \textbf{Why}-questions into a statement so our model can also function as a single-round chitchat chatbot that can answer \textbf{Why}-questions.

%% file: task.tex
\section{Learning to Explain \label{sec:l2e}}






We use the discourse extractor developed by \citet{nie2017dissent}. This extractor first filters sentences that contain a particular discourse marker (in our case, the marker ``because''). It then uses predefined, pattern-based rules on the dependency parse obtained from the Stanford CoreNLP dependency parser \cite{manning_stanford_2014} to split the sentence into two semantically complete sentence clauses, which can be referred as S1 and S2. 
Dependency parsing allows us to isolate explanations and phenomena from exogenous modifying phrases.
Using these patterns to parse sentences with ``because'' also allows us to deal with the free order of the explanation and phenomenon in English.
We formulate the \textbf{L2E} task as: given the phenomenon S1, the model needs to learn to generate a plausible explanation S2.



In addition to retrieving the phenomenon-explanation pair, we additionally retrieve five sentences that immediately precede the phenomenon to provide context. We concatenate the context with S1 using a special separation token, resulting in the sequence \texttt{C1, C2, ..., C5 <SEP> S1}. We hypothesize that context will allow the model to generate more thematically relevant explanations. We refer to this setting as the \textbf{L2EC} task.


At last, we describe a procedure in Algorithm~\ref{alg:s-to-q} that  uses dependency parsing to turn \textbf{Why}-questions into the statement format of S1. This allows us to generate explanations as responses to \textbf{Why}-questions.


\begin{algorithm}[t]
   \caption{Q-to-S1}
   \label{alg:s-to-q}
\begin{algorithmic}
   \STATE {\bfseries Input:} question $q$, dependency parsed.
   \STATE Remove ``Why". Start at the \textsc{root} of $q$: 
   \STATE $subj$ = \textsc{nsubj} or \textsc{nsubjpass} 
   \STATE $aux$ = first dependent in [\textsc{aux}, \textsc{cop}, \textsc{auxpass}] 
   \STATE $vp^{(\text{lemma})}$ = all remaining dependents
   \IF{$aux$ in [``do'', ``does'', ``did'']}
   \STATE $vp$ = apply tense/person of $aux$ to $vp^{(\text{lemma})}$
   \ELSE
   \STATE $vp$ = $aux$ $vp^{(\text{lemma})}$
   \ENDIF
   \STATE $s$ = $subj$ $vp$ 
\end{algorithmic}
\end{algorithm}

%% file: model_new.tex
\section{Model}


\subsection{Language Modeling}

Language modeling focuses on modeling the joint probability of a sequence $p(X=x_1, ..., x_n)$. Using chain rule, this can be decomposed as $p(X)=\prod_{t=1}^n p(x_t|x_{<t})$, the product of conditional probabilities. The model parameterized by $\theta$ optimizes to maximize the log of the likelihood function $\mathcal{L}(X; \theta) = \sum_{t=1}^n p_\theta(x_t|x_{<t})$. In a neural language model, proposed by \citet{bengio2003neural}, a recurrent neural network is trained by truncated backpropogation through time to learn to model (theoretically) an infinitely long sequence.

\subsection{Sequence to Sequence Modeling}

First introduced by \citet{sutskever2014sequence}, sequence-to-sequence (Seq2Seq) modeling estimates a conditional probability distribution of sequence $Y$ given sequence $X$. $p(Y|X)$, where $X = \{x_1, ..., x_n\}$, and $Y = \{y_1, ..., y_k\}$. The overall objective function is similar to a language model: to maximize the log-likelihood of the probability of the $Y$  sequence given the $X$ sequence: $\mathcal{L}(Y, X; \theta, \psi) = \sum_{t=1}^k p_{\theta, \psi}(y_t |y_{<t}, X)$, with parameters $\theta$ for the encoder and $\psi$ for the decoder.
In our work, we experiment with different architectures for the encoder and decoder.




%% file: data.tex
\section{Data}




We provide data accessibility statements in Appendix~\ref{acc:access} for each dataset we use to train and evaluate our models. Our constructed dataset and web demo code are publicly available\footnote{\url{https://github.com/windweller/L2EWeb}}.

\begin{table}[htbp]
\centering
\footnotesize
\begin{tabular}{cccc}
\toprule
 Source Dataset & Task & Data & Length \\
\midrule
 NewsCrawl & L2E & 2.07M & 29.4 \\
 NewsCrawl & L2EC & 2.57M & 149.4  \\
 \midrule
 Winograd & L2E & 61 & 18.0 \\
 COPA & L2E & 250 & 14.2 \\
 News Commentary & L2E/L2EC & 6301 & 28.6 \\
\bottomrule
\end{tabular}
\caption{Top are training datasets and bottom are evaluation datasets for each task. We report the average length of sentences for each dataset (S1 and S2 combined). News Commentary with context has 156.3 words on average.}
 \label{table:description-dataset}
\end{table}

\subsection{Training Data}

\paragraph{NewsCrawl Dataset}

We build up our training dataset from two large news datasets: Gigaword Fifth Edition~\cite{parker2011english} and NewsCrawl ~\cite{bojar2018proceedings}. These two datasets contain news stories from 2001-2017, and are non-overlapping. We built our dataset of News explanation pairs using the pipeline described in Section~\ref{sec:l2e} and then split into training, validation, and test. More details are reported in Appendix~\ref{app:data}.



\begin{table*}
    \centering
    \footnotesize
    \begin{tabular}{l l l c l c}
        \toprule
        Data & S1 & Generated S2 & Rank & Reference S2 & Rank \\
        \midrule
         NewsCrawl & \begin{tabular}{@{}l@{}}That banned his most \\ threatening challenger, \\ Rally leader  Alassane \\ Ouattara, from running for \\ president because $\rule{0.5cm}{0.1mm}$ ? \end{tabular} & He was born in Burkina Faso. & --- & He is only half-Ivorian. & --- \\
        \midrule
         NewsCrawl & \begin{tabular}{@{}l@{}}The victim was only \\ saved because $\rule{0.5cm}{0.1mm}$ ? \end{tabular} & He was wearing a seatbelt. & --- & \begin{tabular}{@{}l@{}} The dog turned on the \\ former lifeguard. \end{tabular} & --- \\
        \midrule
         NewsCrawl & \begin{tabular}{@{}l@{}}I voted for George W. \\ Bush because $\rule{0.5cm}{0.1mm}$ ? \end{tabular} &  \begin{tabular}{@{}l@{}}I thought he was the best \\ person for the job. \end{tabular} & --- & \begin{tabular}{@{}l@{}}That's the name you \\ heard a lot of talk about. \end{tabular} & --- \\
         \midrule
         WSC-G & \begin{tabular}{@{}l@{}}An hour later John left \\ because $\rule{0.5cm}{0.1mm}$? \end{tabular} & He didn't feel safe. & 0.0 & John promised Bill to leave. & 0.67 \\
         \midrule
         COPA & \begin{tabular}{@{}l@{}} The woman gave the \\ man her phone number \\ because $\rule{0.5cm}{0.1mm}$? \end{tabular} &  \begin{tabular}{@{}l@{}}She was too busy to be \\ bothered by the man. \end{tabular} & 0.17 & She was attracted to him. & 0.5 \\
         \midrule
         NC & \begin{tabular}{@{}l@{}} Moreover, ordinary \\ Russians are becoming \\ allergic to liberal  \\ democracy because $\rule{0.5cm}{0.1mm}$? \end{tabular} & \begin{tabular}{@{}l@{}} They see it as a threat \\  to their own interests. \end{tabular} & 0.16 & \begin{tabular}{@{}l@{}} Liberal technocrats have \\ consistently served as \\ window dressing for an \\ illiberal Kremlin regime. \end{tabular} & 0.19 \\
    \bottomrule
    \end{tabular}
    \caption{\textbf{Example pairs} from our highest performing models with the original sentence as a reference. Human ranking score lower is better. We provide examples of especially poor-rated generations in the Appendix.}
    \label{tab:task_examples}
\end{table*}

\paragraph{BookCorpus}
BookCorpus is a set of unpublished novels (\emph{Romance}, \emph{Fantasy}, \emph{Science fiction}, and \emph{Teen} genres) collected by \citet{zhu2015aligning}. We use a publicly available pre-trained BookCorpus language model from \citet{holtzman2018learning}. We refer to this model as \textbf{L2W}.

\paragraph{Language Modeling One Billion} This dataset (LM-1B) is currently the largest standard training dataset for language modeling, roughly the same size as BookCorpus. This dataset is a subset of the NewsCrawl dataset, from 2007-2011. We use a pre-trained language model on this corpus from \citet{jozefowicz2016exploring}. We refer to this model as \textbf{LM-1B}.

\subsection{Evaluation Data}

\paragraph{News Commentary (NC) Dataset}



We collect pairs from a public dataset that contains predominantly commentary written about current news\footnote{https://www.project-syndicate.org/about}.
We use this dataset as the main evaluation of the news-based explanation because 1).~It is a separate dataset without any overlap with NewsCrawl; 2).~This dataset still belongs to the same news domain, so it provides an in-domain evaluation for  \textbf{L2E}, \textbf{L2EC} and \textbf{LM-1B} models.

\paragraph{Winograd Schema Challenge Subset (WSC-G)}

We use 61 example sentences in the Winograd Schema Challenge that contain the words ``because'' or ``so''.
Similar to \citet{trinh2018simple}, we substitute the ambiguous pronouns with the correct referent and ask the model to generate the correct explanation ``the trophy is too big'' to the phenomenon ``The trophy doesn't fit in the suitcase''.



\paragraph{Choice of Plausible Alternatives (COPA)}

\citet{roemmele2011choice} proposed a task that 
contains questions such as ``The women met for coffee. What was the CAUSE of this?'', and the model is asked to choose between two pre-defined causes. In our setting, we directly ask the model to generate a cause. For language models, we append ``because'' to the end of each COPA sentence and ask the model to generate the rest.

%% file: experiments.tex
\section{Experiments}


\subsection{Language Model Training}

We use the same language model described in \citet{holtzman2018learning}. We train 10 epochs for both L2E and L2EC. We use a one layer LSTM~\cite{hochreiter1997long} with 2048 hidden state dimensions and 256 word dimensions. We chose these hyperparameters 
by tuning on the validation set of each task.
Our language model achieved 51.64 perplexity on the L2E test set, and 37.61 perlexity on the L2EC test set.


\subsection{Seq2Seq Model Training}

We experiment with two architectures: LSTM encoder-decoder and Transformer~\cite{vaswani2017attention}. We find that with the L2E task, the Transformer architecture performed better, and for the L2EC task, the LSTM encoder-decoder performed better. We suspect that Transformer is worse when the source sequence is too long.
We tune each architecture's hyperparameters extensively and we pick the best architecture for each task to evaluate on the evaluation datasets.

\begin{table}[htbp]
\centering
\footnotesize
\begin{tabular}{c|c c c c} 
\toprule
 Model & \multicolumn{2}{c}{L2E}  & \multicolumn{2}{c}{L2EC}  \\
 & Acc & Perp & Acc & Perp \\
\midrule
LSTM & 36.2 & 41.4 & \textbf{36.0} & \textbf{41.3} \\ 
Transformer & \textbf{38.2} & \textbf{33.1} & 27.8 & 96.7 \\ 
\bottomrule
\end{tabular}
\caption{We report the best per-token accuracy and perplexity evaluated for each tuned architecture on the L2E/L2EC validation dataset.}
 \label{table:training-perf}
\end{table}

\subsection{Automatic Evaluation}

We use automatic metrics to evaluate the 8 models' performance on the News Commentary dataset. Even though this is a non-overlapping held-out dataset to our news training data, it is still within the same domain. We 
find that L2E/L2EC based models obtained higher scores across all automatic metrics in Table~\ref{table:automatic-performance}.
Our results also demonstrate 
that context matters for explanation.
The L2EC task models, trained on context, can generate higher quality explanations than context-free L2E task models.


\begin{table*}[ht!]
\centering
\footnotesize
\begin{tabular}{c|c c c c c c | c c c} 
\toprule
 Model & \multicolumn{2}{c}{BLEU} & \multicolumn{2}{c}{ROUGE} & \multicolumn{2}{c}{METEOR} & \multicolumn{3}{c}{Human Ranking} \\
 & Greedy & Beam & Greedy & Beam & Greedy & Beam & COPA & WSC-G & NC \\
\midrule
L2E-Seq2Seq & \textbf{0.55} & 0.37 & 18.8 & 18.3 & 7.4 & 7.6 & \textbf{0.412} & \textbf{0.409}  & 0.454 \\
L2EC-Seq2Seq & 0.40 & \textbf{0.47} & \textbf{19.9} & \textbf{19.7} & \textbf{8.6} & \textbf{8.8} & --- & --- & 0.433 \\ 
\midrule
L2E-LM & 0.25 & 0.20 & 15.9 & 16.8 & 6.1 & 6.7 & 0.515 & 0.572 & 0.479  \\
L2EC-LM & 0.36 & 0.38 & 17.0 & 17.7 & 6.7 & 7.3 & --- & --- & \textbf{0.432} \\
\midrule 
LM-1B$^\dagger$ & 0.18 & --- & 16.9 & --- & 7.1 & --- & 0.526 & 0.484 & 0.454 \\
L2W$^\dagger$ & 0.00 & 0.00 & 14.0 & 13.9 & 6.7 & 6.8 & 0.511 & 0.523 & 0.625 \\
L2WC & 0.13 & 0.14 & 12.8 & 12.7 & 5.7 & 5.7 & --- & --- & 0.546 \\
\midrule
OpenSubtitle$^\dagger$ & 0.04 & 0.0 & 13.0 & 13.4 & 1.9 & 3.7 & 0.827 & 0.823 & 0.811\\
\midrule
Reference & 100 & 100 & 100 & 100 & 100 & 100 & \textbf{0.266} & \textbf{0.238} & \textbf{0.267} \\
\bottomrule
\end{tabular}
\caption{BLUE, ROGUE, METEOR are evaluated on News Commentary test data. Any model with \textbf{C} in the name is evaluated with full context. Models with $^\dagger$ are pre-trained models from other work. Only L2E-Seq2Seq uses the Transformer architecture, the rest LSTM. In human ranking, we report the average rank across participants. Top ranking is 0 and lowest ranking is 1. 
}
 \label{table:automatic-performance}
\end{table*}


\subsection{Human Evaluation}

\paragraph{Ranking Explanations}

We evaluate the models' relative performance on generating explanations through a survey with human evaluators.
75 participants were recruited using Amazon's Mechanical Turk (AMT).
Each evaluator saw 10 prompts from a single dataset, and ranked 7 to 9 explanations: the original explanation extracted from the dataset and the explanations generated by different models.
30 participants saw prompts from our Winograd dataset, 30 participants saw prompts from News Commentary, and 15 participants saw prompts from COPA.
We report the results of this evaluation in the Human Ranking subsection of Table \ref{table:automatic-performance}.

\paragraph{Rating Explanations}

\begin{table}[t]
\centering
\footnotesize
\begin{tabular}{r|c c} 
\toprule
& Original & L2E-Seq2Seq \\
\midrule
Goodness & \textbf{0.699} [0.67, 0.72] & 0.500 [0.45, 0.55] \\	
Relatedness & \textbf{0.723} [0.70, 0.74] & 0.590 [0.55, 0.63] \\
Grammaticality & 0.684 [0.66, 0.71] & \textbf{0.738} [0.70, 0.77] \\
Helpfulness & \textbf{0.696} [0.67, 0.72] & 0.512 [0.47, 0.56] \\
Plausibility & \textbf{0.710} [0.69, 0.73] & 0.543 [0.50, 0.59] \\
\bottomrule
\end{tabular}
\caption{Results of rating study with human evaluators, average rating and bootstrapped 95\% CI.}
 \label{table:ratings}
\end{table}

In a followup survey, 60 human evaluators on AMT rated explanations generated by the L2E-Seq2Seq model with beam search and the original (between participants). Ratings were from 0 (extremely bad) to 1 (extremely good) along various dimensions of explanation quality. Results of this study are shown in Table \ref{table:ratings}. Generated explanations overall were rated worse than human explanations, but tended to be more good than bad ($\geq 0.5$) on all measures.

%% file: discussion.tex
\section{Discussion}


The nature of phenomenon-explanation mapping has always been one-to-many. People can offer drastically different explanations to the same phenomenon. We argue that requiring the machine to generate plausible explanations is more useful and therefore a better goal for models to achieve.
Models trained on traditional chatbot corpora are unable to answer why questions because of data sparsity.
We note that the generated results are not similar to the original explanations but are often acceptable by human assessment.

\paragraph{Features of Explanations}
In the human rating experiment, our model was overall rated higher than the original explanations only on the grammaticality measure. However, this measure seems least representative of the overall explanation quality: ratings for most features were highly correlated with each other (0.771-0.865), but not with grammaticality (0.196-0.323). This shows that, while we can achieve plausible explanations with our models, more research is required in order to reach human-level quality.

\paragraph{Explaining as Generating} 
Even though formulating the task of providing explanation as a sequence generation task allows us to leverage the rapid advancements in the natural language generation community, we sidestep a vast amount of literature that aims to provide informatively \emph{correct} explanations as well as grounding explanations theoretically to the causal understanding of the situation~\cite{halpern2005causes}. We also suffer from the same drawbacks noticed in natural language generation papers such as brevity and generic responses, failure to leverage long context, and being data hungry~\cite{holtzman2018learning}.

\paragraph{Exploring Linguistic Structures} 
The curated dataset of explanation-phenomenon pairs provides an opportunity to explore descriptive structures and features of explanations. In principle, one can use this dataset to formulate frequent and common syntactic and semantic patterns for natural-sounding explanations.  This would aid our understanding of how why-questions can be addressed satisfactorily.






%% file: conclusion.tex
\section{Conclusion}

We present the task of generating plausible explanations as an important goal for neural sequence-to-sequence models.
We curate a large dataset of phenomenon-explanation pairs so that these models can learn to provide plausible explanations as judged by humans, and formulate responses to general \textbf{Why}-questions.

\section*{Acknowledgement}

We thank Barry Haddow and Michael Hahn for their advice and assistance.
The research is based upon work supported by the Defense Advanced Research Projects Agency (DARPA), via the Air Force Research Laboratory (AFRL, Grant No. FA8650-18-C-7826). The views and conclusions contained herein are those of the authors and should not be interpreted as necessarily representing the official policies or endorsements, either expressed or implied, of DARPA, the AFRL or the U.S. Government.
The U.S. Government is authorized to reproduce and distribute reprints for Governmental purposes notwithstanding any copyright annotation thereon.

%% file: appendix.tex
\section{Supplementary Materials}

\begin{table*}
    \centering
    \footnotesize
    \begin{tabular}{l l l c l c}
        \toprule
        Data & S1 & Generated S2 & Rank & Reference S2 & Rank \\
        \midrule
         NewsCrawl & \begin{tabular}{@{}l@{}} Perhaps the student was \\ so good because $\rule{0.5cm}{0.1mm}$ ? \end{tabular} & He was so good at it. & --- & \begin{tabular}{@{}l@{}}Thursday wasn't the first \\ time he acted as a moderator. \end{tabular} & --- \\
        \midrule
         NewsCrawl & \begin{tabular}{@{}l@{}} Cheaper prices drove \\ computers out of the walls \\ of these big companies \\ because $\rule{0.5cm}{0.1mm}$ ? \end{tabular} & \begin{tabular}{@{}l@{}} They were cheaper than \\ the ones that were being \\ used. \end{tabular} & --- & \begin{tabular}{@{}l@{}} You suddenly didn't need \\ all that money anymore. \end{tabular} & --- \\
        \midrule
         WSC-G & \begin{tabular}{@{}l@{}} Sam and Amy are \\ passionately in love, \\ but Amy's parents are  \\ unhappy about it \\ because $\rule{0.5cm}{0.1mm}$? \end{tabular} & \begin{tabular}{@{}l@{}} They don't want to be \\ the first female president. \end{tabular} & 0.87 & Sam and Amy are fifteen. & 0.37 \\
         \midrule
         COPA & \begin{tabular}{@{}l@{}} The hamburger meat \\ browned because $\rule{0.5cm}{0.1mm}$? \end{tabular} &  \begin{tabular}{@{}l@{}} That's what it is. \end{tabular} & 0.83 & The cook grilled it. & 0.0 \\
         \midrule
         NC & \begin{tabular}{@{}l@{}} The desperately poor may \\ accept handouts because $\rule{0.5cm}{0.1mm}$? \end{tabular} & \begin{tabular}{@{}l@{}} They are the only ones \\ who can afford it. \end{tabular} & 0.83 & \begin{tabular}{@{}l@{}} They feel they have to. \end{tabular} & 0.12 \\
    \bottomrule
    \end{tabular}
    \caption{\textbf{Bad example pairs} from our lowest performing models with the original sentence as a reference. Human ranking score lower is better. Full list of WSC-G and COPA generations can be found in \url{https://github.com/windweller/L2EWeb/blob/master/WinogradS2Generation.ipynb}.}
    \label{tab:bad_examples}
\end{table*}

\subsection{Data Accessibility Statement \label{acc:access}}
The majority of the data we use are publicly available. We provide specific instructions on how to obtain these data below:

\noindent\textbf{Gigaword 5th Edition} This dataset is provided through Linguistic Data Consortium (LDC): \url{https://catalog.ldc.upenn.edu/LDC2011T07}. Even though this dataset is only available through subscription, most university libraries should have existing subscriptions, and only 20\% of our training data comes from this dataset.

\noindent\textbf{News Crawl Dataset} The shuffled version of this dataset is publicly available\footnote{\url{http://www.statmt.org/wmt18/translation-task.html}}. We requested the original un-shuffled dataset from Barry Haddow\footnote{\url{http://homepages.inf.ed.ac.uk/bhaddow/}} so that we can extract context for L2EC task. We believe this dataset can be easily accessed by the public upon an email request.

\noindent\textbf{BookCorpus} This dataset is no longer publicly available. However, there are many neural language models pre-trained on this dataset that are publicly available. We used one that can be accessed from \url{https://github.com/ari-holtzman/l2w}.

\noindent\textbf{News Commentary Dataset} This is also publicly available through the WMT workshop\footnote{\url{http://data.statmt.org/wmt18/translation-task/news-commentary-v13.en.gz}} similar to the NewsCrawl dataset. This dataset is not shuffled.

\noindent\textbf{Winograd Schema Challenge} The original version of this dataset is publicly available \url{https://cs.nyu.edu/davise/papers/WinogradSchemas/WS.html}. We use a processed version from \citet{trinh2018simple}, which can be accessed through Google Cloud Storage: \url{gs://commonsense-reasoning/reproduce/commonsense_test/wsc273.json}.

\noindent\textbf{Choice of Plausible Alternatives} This dataset is available at \url{http://people.ict.usc.edu/~gordon/copa.html}.

\subsection{Training Data Curation \label{app:data}}

In order to automatically curate a sizable amount of training data, we choose large corpora that are made of news articles, due to the well-formedness of sentences and there are many phenomenon-explanation pairs in news stories. We use Gigaword fifth edition~\cite{parker2011english} which contains news stories from seven news agencies over the span of 2001-2010. We extracted paragraphs and tokenized the sentences. We discard non-English characters. Another large dataset of new articles comes from WMT-18, the NewsCrawl dataset~\cite{bojar2018proceedings}. This dataset spans from 2007-2017 collected from the RSS (Rich Site Summary) feed of 18 news agencies. The only overlapping agency between Gigaword and NewsCrawl is Los Angeles Times. In addition to the randomly shuffled dataset we obtained from the WMT-18 website, we additionally contacted the organization for the unshuffled version of data. We refer to this dataset as the NewsCrawl-ordered. This dataset is slightly larger than the current released version of NewsCrawl and contains a couple of months of early 2018 data. We shuffle and then split both datasets into train/valid/test in standard 0.9/0.05/0.05. We use the validation and test set on this task to pick the best performing model.

\subsection{Language Model Details}
We use adaptive gradient descent (AdaGrad) with learning rate 0.1 and weight decay of 1e-6.

\subsection{Seq2Seq Model Details}

We built and trained our Seq2Seq model using OpenNMT~\cite{2017opennmt}. For the L2E task, we used a 6-layer Transformer model, with hidden dimension 512, feedforward layer dimension 2048, and 8 attention heads. We train with dropout rate of 0.1 with Noam optimizer.
For the L2EC task, we used a 2-layer LSTM model with 650 hidden dimension size for both encoder and decoder, as well as for word embedding. We train with dropout rate of 0.2 and Noam optimizer.


\begin{figure}[ht]
\centering
\includegraphics[trim=2cm 7cm 0cm 7cm, width=\columnwidth]{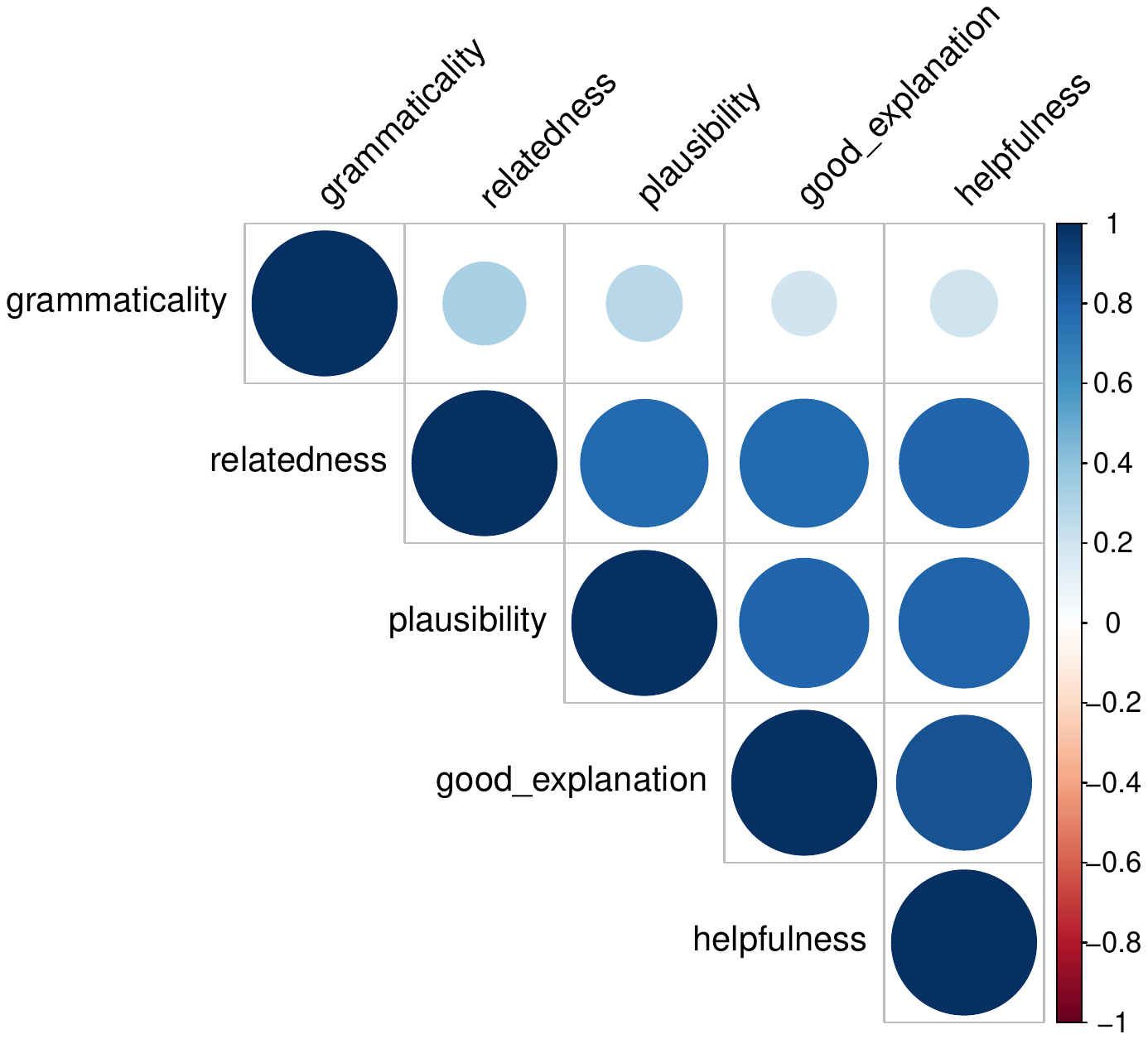}
\caption{Correlations of human ratings on Winograd Schema Challenge explanations}
 \label{fig:cor}
\end{figure}

\begin{figure}[ht]
\centering
\includegraphics[trim=2cm 0cm 0cm 0cm, width=\columnwidth]{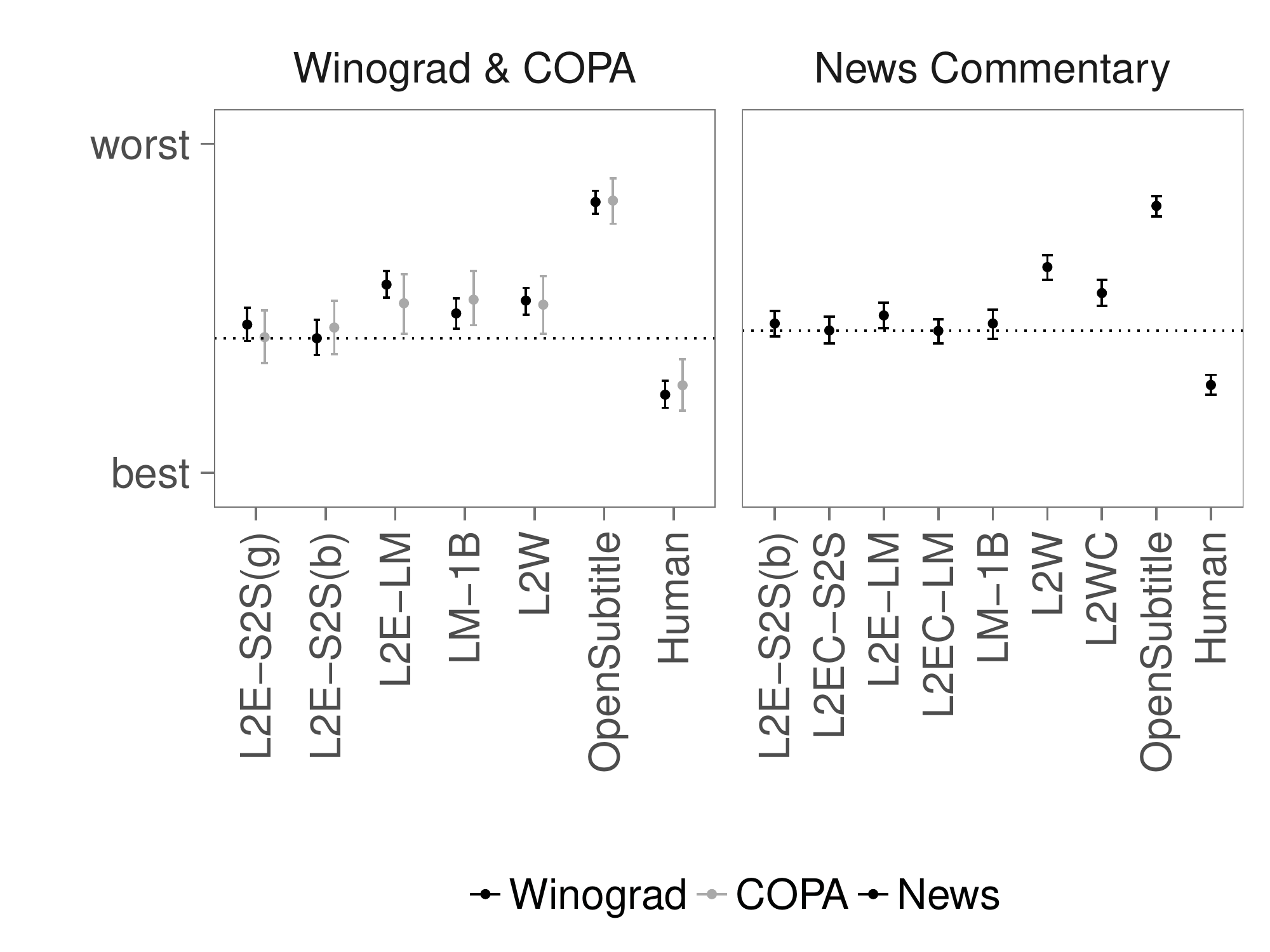}
\caption{The average ranking of each model's generated response (lower is better).}
 \label{fig:cor}
\end{figure}

\begin{figure}[ht]
\centering
\includegraphics[trim=2cm 0cm 0cm 0cm, width=\columnwidth]{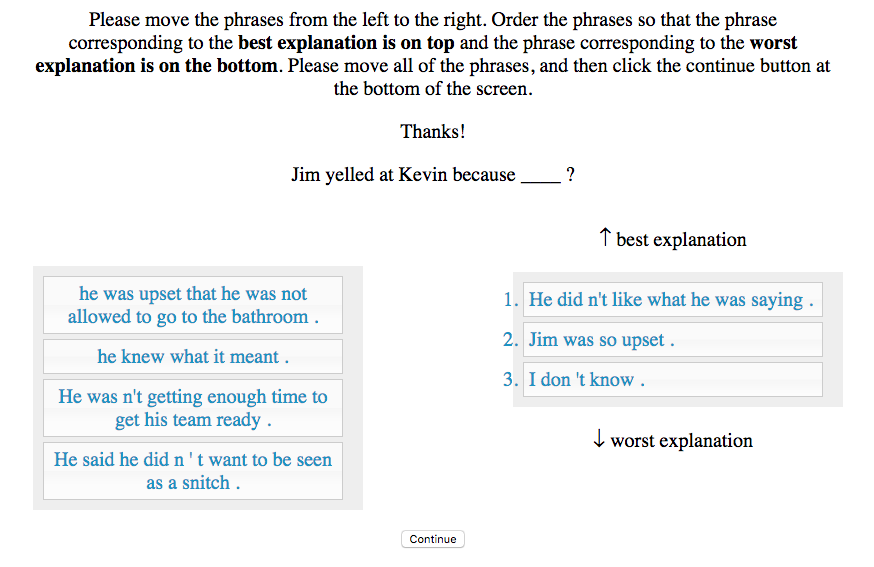}
\caption{Screenshot of raking study.}
 \label{fig:rankings}
\end{figure}

\begin{figure}[ht]
\centering
\includegraphics[trim=2cm 0cm 0cm 0cm, width=\columnwidth]{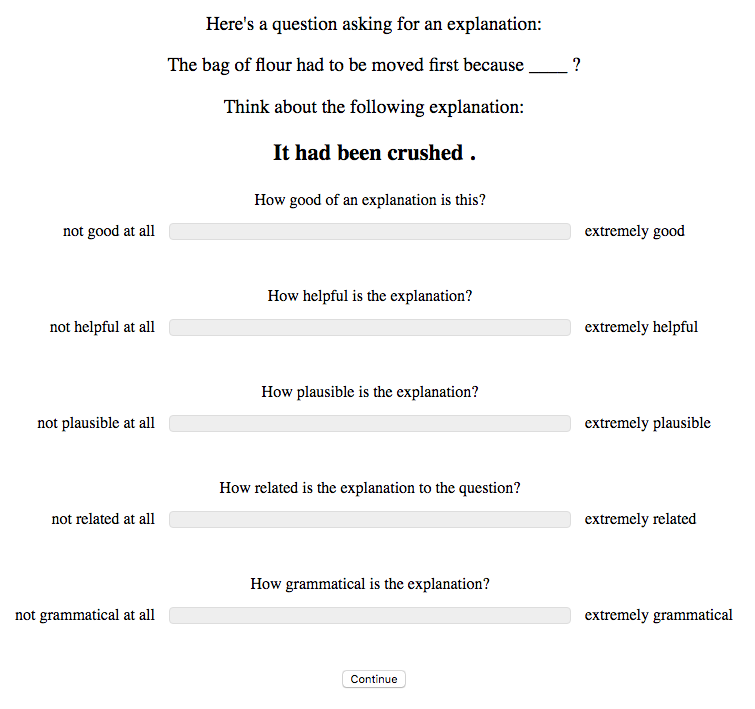}
\caption{Screenshot of ratings study.}
 \label{fig:ratings}
\end{figure}